\documentclass[11pt]{article}

\usepackage[preprint]{acl}

\usepackage{times}
\usepackage{latexsym}

\usepackage[T1]{fontenc}

\usepackage[utf8]{inputenc}

\usepackage{microtype}

\usepackage{graphicx}

\usepackage{booktabs}

\title{GAIA-v2-LILT: Multilingual Adaptation of Agent Benchmark\\beyond Translation}

\author{\hspace{0.5cm} Yunsu Kim$^{*}$ \hspace{1em} Kaden Uhlig$^{*}$ \hspace{1em} Joern Wuebker \\
  LILT \\
\texttt{\{yunsu.kim,kaden.uhlig,joern\}@lilt.com}}

\begin{document}

\maketitle
\def\thefootnote{*}\footnotetext{Equal contribution.}\def\thefootnote{\arabic{footnote}}

\begin{abstract}
  Agent benchmarks remain largely English-centric, while their multilingual versions are often built with machine translation (MT) and limited post-editing.
  We argue that, for agentic tasks, this minimal workflow can easily break benchmark validity through query-answer misalignment or culturally off-target context.
  We propose a refined workflow for adapting English benchmarks into multiple languages with explicit functional alignment, cultural alignment, and difficulty calibration using both automated checks and human review.
  Using this workflow, we introduce \textit{GAIA-v2-LILT}, a re-audited multilingual extension of GAIA covering five non-English languages.
  In experiments, our workflow improves agent success rates by up to 32.7\% over minimally translated versions, bringing the closest audited setting to within 3.1\% of English performance while substantial gaps remain in many other cases.
  This indicates that a substantial share of the multilingual performance gap is benchmark-induced measurement error, motivating task-level alignment when adapting English benchmarks across languages.
  The data is available as part of the MAPS package.\footnote{\url{https://huggingface.co/datasets/Fujitsu-FRE/MAPS/viewer/GAIA-v2-LILT}} We also release the code used in our experiments.\footnote{\url{https://github.com/lilt/gaia-v2-lilt}}

\end{abstract}

\section{Introduction}
Modern AI systems are moving from single-turn assistants to autonomous agents that perform multi-step reasoning with external tools \cite{mialon2023augmented,wang2024autonomousagentssurvey,xi2025risepotential}.
However, agent benchmarks remain largely English-centric \cite{yao2022webshop,deng2023mind2web,liu2024agentbench,zhou2024webarena,jimenez2024swebench,mialon2023gaia,barres2025tau2bench,wei2025browsecomp,xie2024osworld,patwardhan2025gdpval}.
This limits reliable measurement for non-English users and reinforces an English bias in training and optimization of a large language model (LLM) \cite{ahuja2023mega,zhang2023donttrust,nguyen2024culturax}.

A common workaround is to machine-translate an English dataset and apply basic post-editing to a limited subset \cite{hofman2025maps,wang2025xwebagentbench,issaka2026translationproxy}.
This approach is standard for conventional document translation, but in agent benchmarks it can fail to preserve task mechanics.
In Figure 1, the answer was incorrectly translated into the Arabic word for ``lion cub'', interpreting the IOC country code ``CUB'' (for Cuba) as a common noun.
Not only is the translation semantically irrelevant, but it also violates the query's constraint to provide a three-letter country code.

\begin{figure}[t]
  \centering
  \includegraphics[width=\columnwidth]{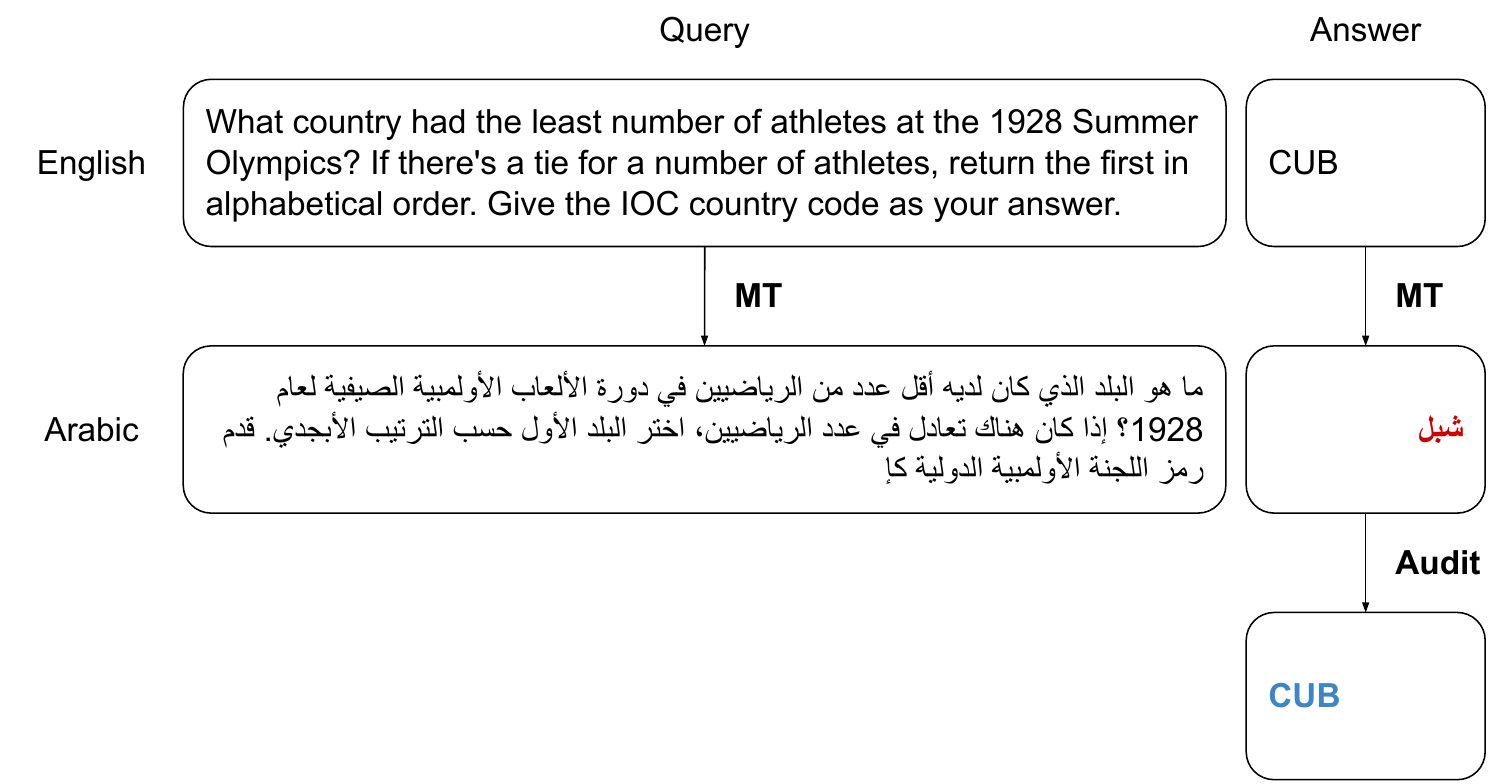}
  \caption{Example of query-answer misalignment by over-translation. MT violates the required answer format, which is subsequently corrected during the human audit.}
  \label{fig:intro:example}
\end{figure}

Including such functional misalignment, this paper analyzes common pitfalls in translation-based workflows for adapting English agent benchmarks into other languages and proposes practical countermeasures.
Our contributions are as follows:
\begin{itemize}\itemsep0em
  \item We show how conventional translation quality issues can compromise the integrity of agent tasks.
  \item We identify specific issues that affects the validity and appropriateness of multilingual agent benchmarking: functional alignment, cultural alignment, and difficulty calibration.
  \item We propose a refined workflow for addressing these issues, combining automatic checks and targeted human review.
  \item We apply this workflow to build the GAIA-v2-LILT dataset and show that correcting these issues materially changes measured agent performance.
\end{itemize}

\section{Related Work}
\paragraph{Multilingual Agent Benchmarks}
Existing multilingual agent benchmarks span general reasoning \cite{hofman2025maps}, function calling \cite{kulkarni2025massiveagents, luo2026lost}, coding \cite{raihan2025mhumaneval}, web navigation \cite{wang2025xwebagentbench}, and computer use \cite{yang2025macosworld}.
They rely on MT from English with minimal or no human edits, whereas our work introduces dedicated alignment procedures tailored for agentic tasks.

Some recent works build multilingual agent tasks from scratch using native components \cite{almeida2025ticket,kautsar2025seadialogues,zhou2025browsecomp}.
While ideal for capturing localized nuances, native creation is costly and labor-intensive to set up for each language.
This work focuses on adapting well-established English benchmarks for standardized evaluation and scalability.

\paragraph{MT Post-Editing}
Conventional MT post-editing \cite{balling2014post,koponen2016machine,vieira2019post} primarily prioritizes fluency and adequacy of the translated text, usually based on established error types \cite{lommel2013multidimensional}.
Recent studies show that reviewers need additional context around the translation for fixing document-level errors, such as inconsistent terms or unclear pronouns \cite{agrawal2024assessing, koneru2024contextual,castilho2025survey}.

While these methods focus on linguistic correctness and meaning preservation, our approach goes beyond surface-level language fixes to ensure functional alignment, cultural relevance, and consistent task difficulty so the benchmarks remain executable and solvable.

\section{Task Definition}
\label{sec:task}
We explain the basics of agentic problem solving to clarify our data curation and benchmarking.

\paragraph{Task}
In this paper, an agentic task is a user query that requires at least one external \textit{action} beyond internal language modeling.
Typical actions include web search, page parsing, file inspection, numerical computation, and scripting.
A successful agent coordinates these steps to return an answer matching the exact output specification.
This differs from standard question answering because the model must plan and execute a procedure rather than just write a fluent response.

\paragraph{Properties}
Following the GAIA dataset \cite{mialon2023gaia}, our tasks feature:
\begin{itemize}\itemsep0em
  \item Single-turn interactions.
  \item Queries often have strict formatting constraints for the output, such as digit precision or list order.
  \item Text-based expected answers to enable deterministic grading.
\end{itemize}

\paragraph{Solving Process}
Agent's execution involves four stages: query comprehension, plan construction, tool execution, and response synthesis.
Agents allocate a finite budget across these steps; If a query is malformed or culturally misaligned, the agent might waste this budget fixing instructions instead of solving the task.

\paragraph{Evaluation}
Most setups measure accuracy on the final answer using an exact match.
To focus on reasoning, we allow minor variations by removing spaces, lowercasing, and stripping punctuation \cite{mialon2023gaia}.
While practical, this binary metric ignores the intermediate steps; an agent might reason correctly but fail due to a flawed answer key or an ill-formed query.

\section{Multilingual Adaptation}
\label{sec:multi}

We explain how to adapt an English task (from Section \ref{sec:task}) into another language.
For each step, we detail the motivation and provide examples of common issues.

\subsection{Translation}
\label{sec:multi:trs}
The first step translates the English data into the target language.
This aims to maintain the dataset size and topic coverage while ensuring the invariance of task function across languages.
Typically, MT performs the first pass, but it introduces two main classes of problems.

\paragraph{Translationese}
Linguistic artifacts arise from literal translations by MT that create \textit{translationese}, which deviates from natural patterns of the target language.
It typically manifests through retaining word order of the source language, word-for-word conversion of idioms, incorrect formality or honorifics \cite{koppel2011translationese,li2025lost}.

\begin{figure}[t]
  \centering
  \includegraphics[width=\columnwidth]{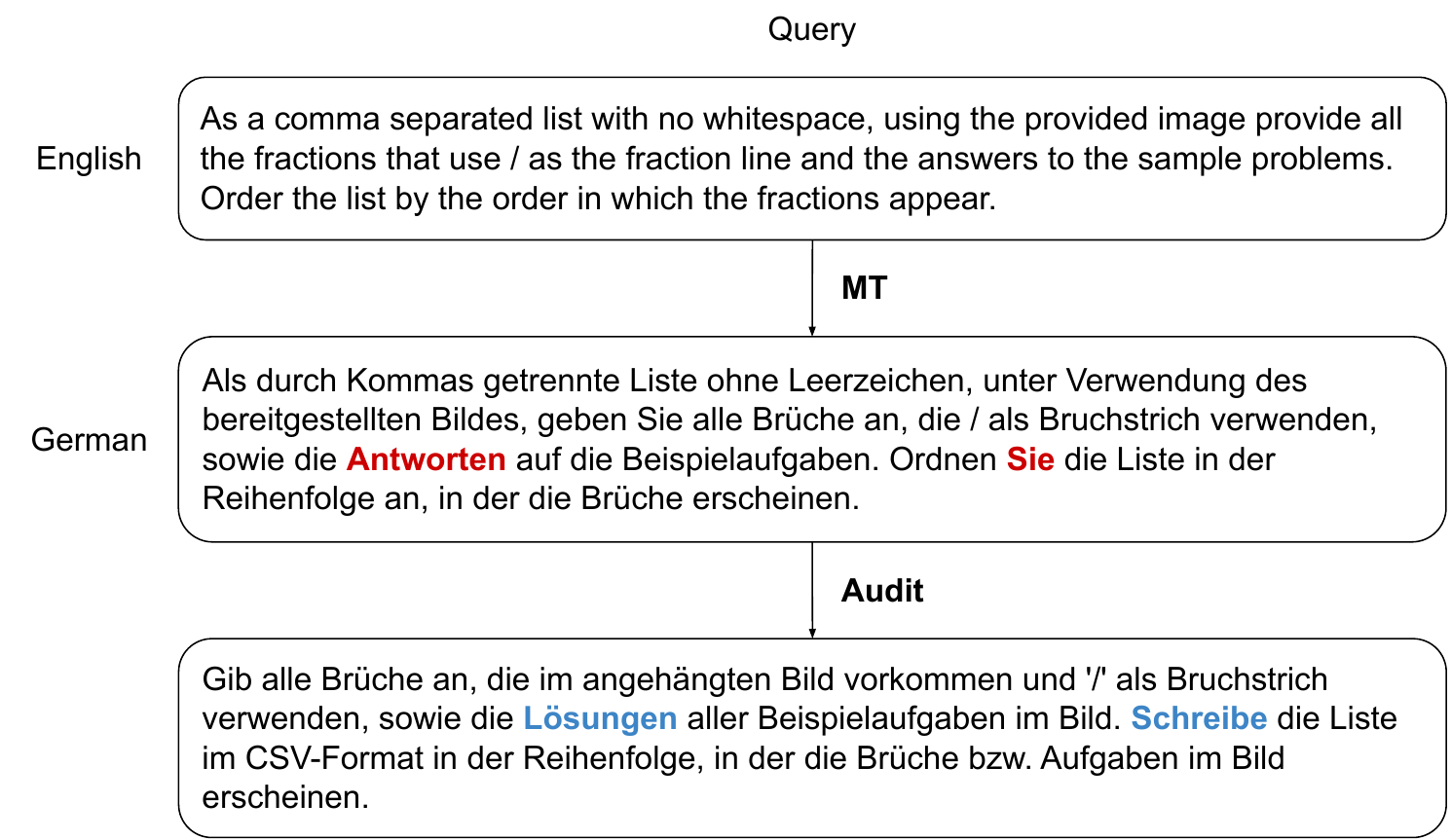}
  \caption{Example of correcting translationese. The audit refines the literal translation to ensure natural phrasing, correct terminology, and appropriate formality.}
  \label{fig:multi:trs:trsltn}
\end{figure}

The resulting data carries an inherently artificial quality, as shown in the example in Figure \ref{fig:multi:trs:trsltn}.
Here, \textit{Antworten} is used for mathematical problems where \textit{Lösungen} is the standard term.
Also, it uses the formal pronoun \textit{Sie}, which is inappropriate for usual human-AI interaction.
The corrected text ensures task realism through native syntax, proper formality, and idiomatic phrasing.

These patterns are especially prominent in modern models trained on vast amounts of synthetic data.
In an agentic context, this forces the model to decode unnatural phrasing rather than focus on high-level reasoning, compromising its planning and execution logic \cite{hofman2025maps,wang2025xwebagentbench}.

\paragraph{Hallucination}
The MT field is transitioning from neural machine translation (NMT) to LLMs for improved contextual fluency.
While LLMs produce more natural prose, they may introduce generative artifacts beyond linguistic errors such as outputting a completely wrong language, including a hint or the gold answer in the prompt \cite{huang2025survey}.

\begin{figure}[t]
  \centering
  \includegraphics[width=\columnwidth]{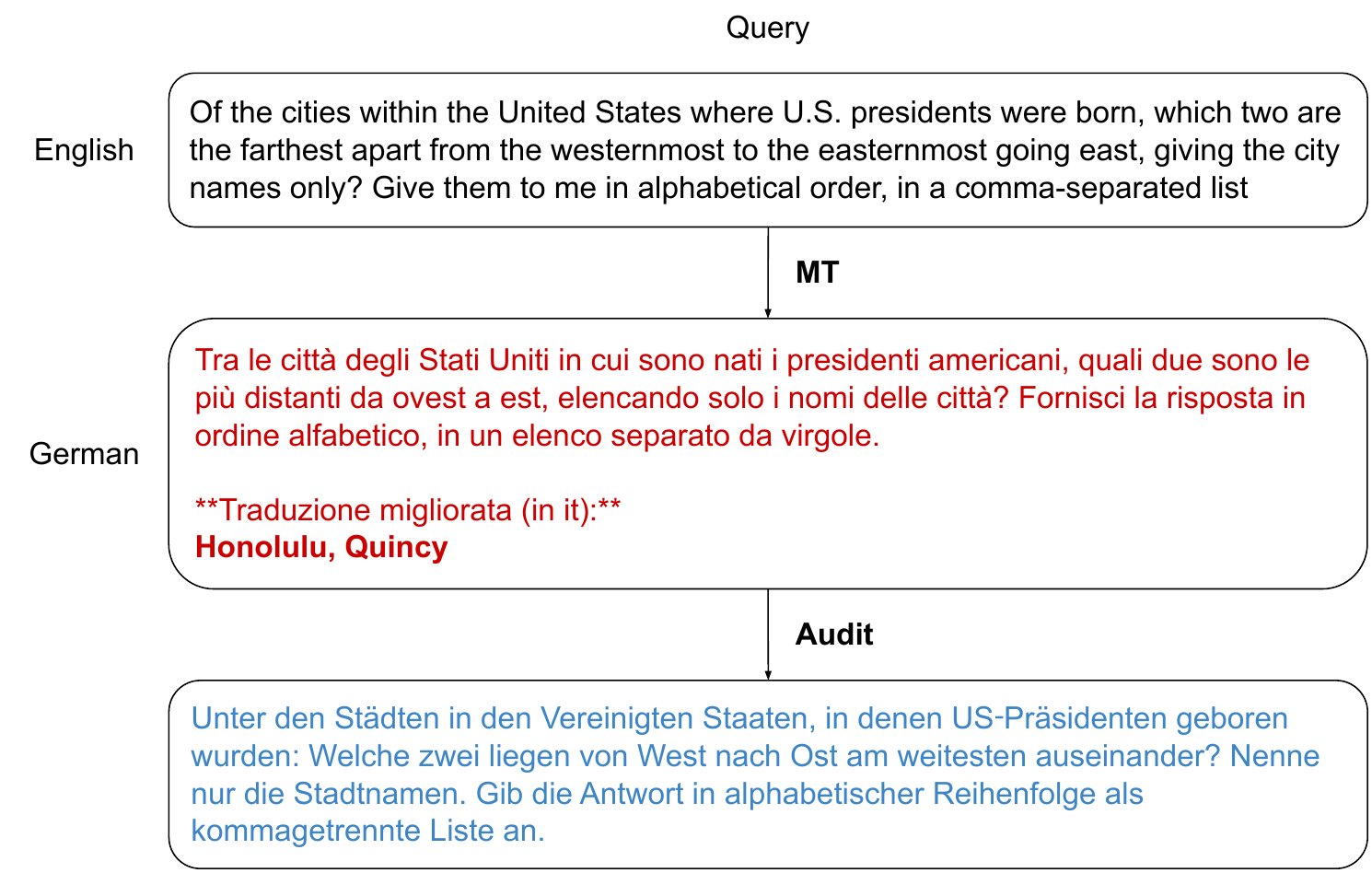}
  \caption{Example of correcting hallucination. Instead of the target German, the MT generates Italian and hallucinates the actual answer within the text. The audit restores the correct language and removes the answer.}
  \label{fig:multi:trs:hallu}
\end{figure}

Such artifacts cause execution failures or shortcuts, shifting the benchmark from a test of intelligence to a test of robustness against corrupted logic.
In Figure \ref{fig:multi:trs:hallu}, the model generated text in Italian rather than the requested German.
Furthermore, it answered the prompt during translation (“\textit{Honolulu, Quincy}”).

\subsection{Functional Alignment}
\label{sec:multi:func}
While standard post-editing can fix the translation issues above, it does not guarantee task solvability.
A critical failure mode is query-answer mismatch, where the translated answer no longer aligns with the translated prompt.
We divide this into two cases:

\paragraph{Under-Translation}
This occurs when a reference answer remains in English even though the context requires a localized term, causing false negatives.
If an agent outputs the correct local name but the evaluator expects the English original, valid reasoning is unfairly penalized.
For example, a German chess task will incorrectly reject the valid local notation \textit{Td5} (``Turm'') if the answer key is left as the English \textit{Rd5} (``Rook'').

\paragraph{Over-Translation}
Conversely, some answers must remain in English to maintain task integrity.
If a task requires extracting a specific string from an English document, translating that string prevents exact matching (Figure \ref{fig:intro:example}).
Similarly, transliterating obscure entities that lack established local conventions often introduces extra ambiguity.

\medskip\noindent
Distinguishing between these cases is a delicate nuance problem.
It requires reviewers to balance deep cultural conventions with a technical understanding of the question type to determine whether localization aids or obstructs the evaluation logic.

\subsection{Cultural Alignment}
\label{sec:multi:cult}
Even when tasks remain solvable, retaining assumptions in the source region can make them unnatural for local users.
These include geography, policy logic, measurement units, platform conventions, and social norms.

In Figure \ref{fig:multi:cult:example}, a query about returning bottles in a road trip was translated into Korean but remained anchored to U.S. highways and imperial units.
The corrected version replaced route references, converted units, updated the date context, and switched to a Korean recycling incentive system (where no local equivalent exists, the literal U.S. context is retained).

While literal translations represent realistic scenarios for expats, full localization better reflects the authentic usage of the primary target population.
Crucially, leaving queries as literal translations often prompts agents to simply translate the text back to English and solve the original U.S.-centric task, bypassing true multilingual reasoning \cite{li2025language,wang2025lost}.

\begin{figure}[t]
  \centering
  \includegraphics[width=\columnwidth]{figures/korean-cultural.pdf}
  \caption{Example of cultural alignment. The MT retains U.S. geography and imperial units. The audit localizes the query with Korean routes, measurements, and a reward system. The date is also updated to align with the actual availability of recycling rewards in South Korea.}
  \label{fig:multi:cult:example}
\end{figure}

\subsection{Difficulty Calibration}
\label{sec:multi:diff}
Beyond linguistic, functional, and cultural alignment, converting a task can still alter its inherent difficulty.
For instance, a query becomes significantly harder if relevant information is scarce on the web in the target language, or if localized currencies complicate simple math.
Therefore, reviewers must manually verify the solving process to ensure the difficulty aligns with the English original, e.g., by conducting local web searches themselves.

\begin{figure}[!h]
  \centering
  \includegraphics[width=\columnwidth]{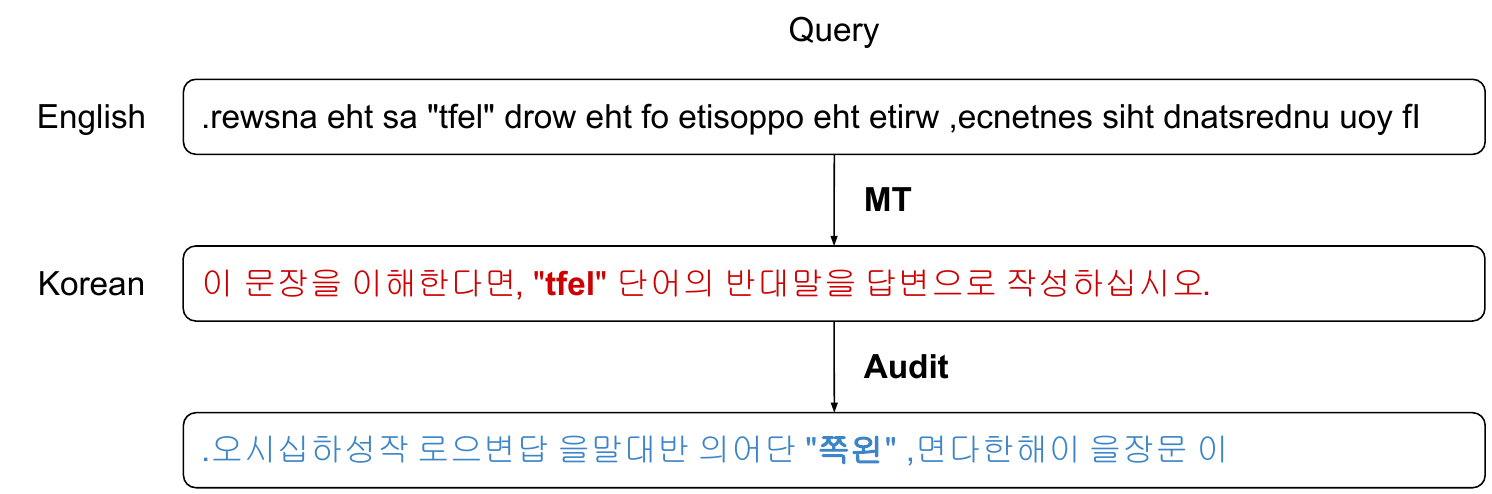}
  \caption{Example of difficulty calibration.}
  \label{fig:multi:diff:example}
\end{figure}

Figure \ref{fig:multi:diff:example} demonstrates difficulty calibration using a word riddle.
While the draft is fluent and preserves the main task function, it lowers the reading difficulty by writing the sentence normally around the English clue ``\textit{tfel}''.
The corrected version restores the original challenge by fully reversing the Korean sentence and localizing the clue word.

\section{GAIA-v2-LILT}
\label{sec:data}
Considering all dimensions of Section \ref{sec:multi}, we constructed a multilingual agent benchmark \textit{GAIA-v2-LILT}, based on the machine-translated version of GAIA \cite{mialon2023gaia} by MAPS \cite{hofman2025maps}.
GAIA-v2-LILT contains 165 query-answer pairs (validation set) for each of the five non-English languages: Arabic, German, Hindi, Korean, and Portuguese (Brazil).

\begin{figure*}[t]
  \centering
  \includegraphics[width=\textwidth]{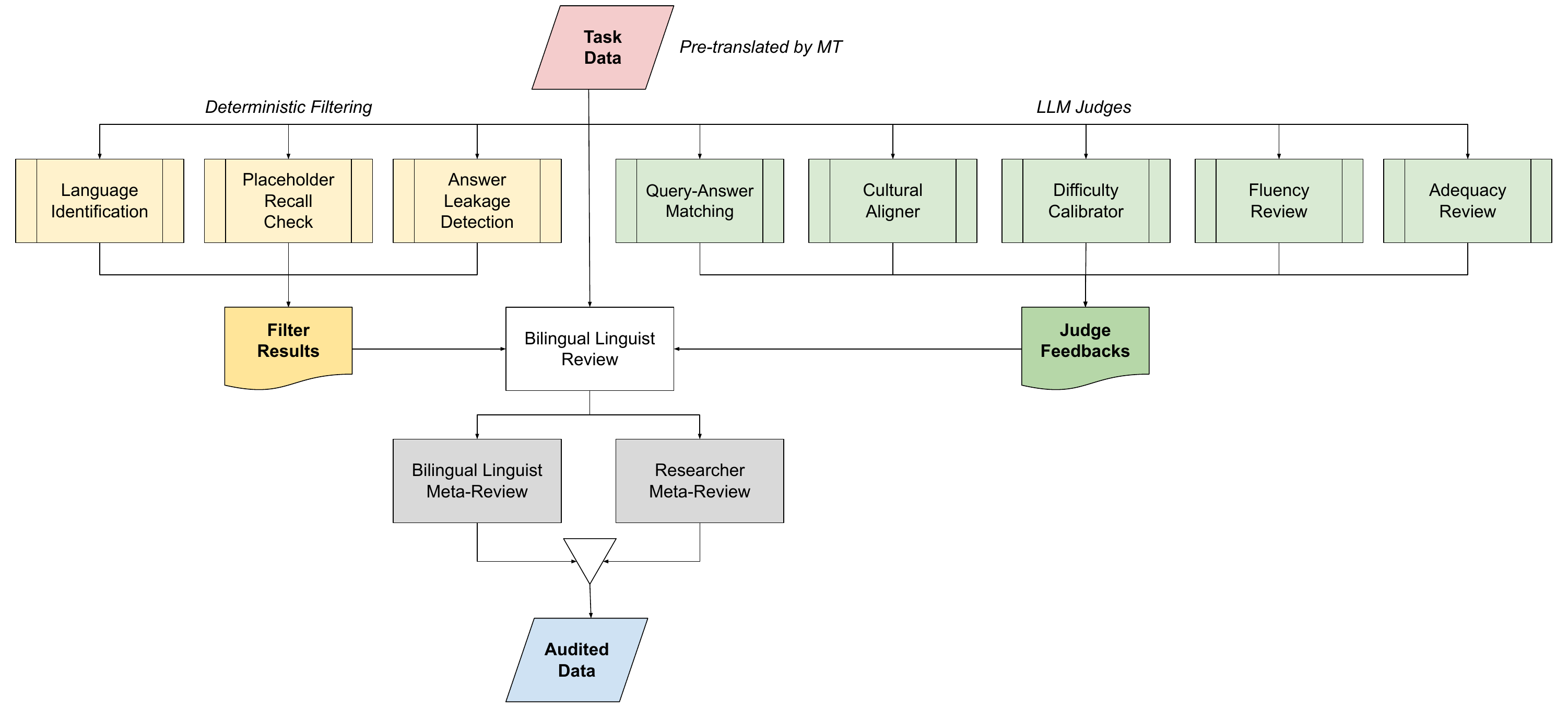}
  \caption{Review workflow for constructing GAIA-v2-LILT.}
  \label{fig:data:workflow}
\end{figure*}

We built a hybrid review workflow combining automatic checks with specialized human auditing to effectively correct the aforementioned issues (Figure \ref{fig:data:workflow}).
It is specifically designed to counter two major evaluation pitfalls:
\begin{itemize}\itemsep0em
  \item LLM's \textit{self-preference}: models favoring text similar to their own outputs \cite{zheng2023judging}
  \item Human's \textit{fluency bias}: reviewers mistakenly assuming that well-written text is technically correct \cite{gudibande2024false}
\end{itemize}

\paragraph{Deterministic Filtering}
Before applying model or human judgment, we used fast, rule-based scripts to catch high-impact, objective defects.
Specifically, these scripts check:
\begin{itemize}\itemsep0em
  \item whether the translation is in the target language (language identification)
  \item whether the expected answer is leaked in the query (string matching)
  \item whether all fixed term categories are preserved, e.g., numbers, URLs, etc. (placeholder recall)
\end{itemize}
Relying on exact matching rather than generative evaluation, this step does not have the self-preference problem of LLMs.

\paragraph{LLM Judges}
Next, we used LLMs to pre-evaluate the deeper semantic and functional aspects.
Rather than asking for a single holistic score, we decomposed the review into narrow, single-axis judgments.
This design helps limit self-preference bias and can better track human judgments, because each decision is made over a short context and a concrete question instead of requiring a deep global analysis of the text \cite{saha2024branch,feng2025m,lee2025checkeval}.
Accordingly, separate judges independently evaluated fluency, adequacy, query-answer compatibility, and cultural appropriateness using binary labels.

\begin{table}[!b]
  \centering
  \begin{tabular}{lccc}
    \toprule
    \textbf{Language} & \textbf{Task} & \textbf{Word} & \textbf{Char} \\
    \midrule
    Arabic & \enspace84.8 & 25.4 & 19.4 \\
    German & \enspace81.2 & 30.0 & 22.2 \\
    Hindi & 100.0 & 55.4 & 46.3 \\
    Korean & \enspace92.1 & 36.9 & 27.9 \\
    Portuguese & \enspace 87.9 & 25.0 & 19.5 \\
    \bottomrule
  \end{tabular}
  \caption{Edit rates [\%] of GAIA-v2-LILT against the original MAPS version of GAIA.}
  \label{tab:data:edit}
\end{table}

\paragraph{Human Audit}
Finally, we employed bilingual human linguists to review the pre-translated tasks.
We conducted 1-on-1 training calls with each reviewer to detail the alignment issues from Section \ref{sec:multi}, placing strict emphasis on validating functional integrity over mere fluency.

During the audit, results from the deterministic filters and LLM judges were displayed alongside the task text.
This highlights critical issues immediately, helping reviewers prioritize their time budget toward the most difficult cases.
Reviewers also had to mark explicit checkboxes for whether each issue category applied to the task.
These structured steps help reduce human fluency bias.

Each task was reviewed by one linguist and meta-reviewed by another linguist and a machine learning researcher.
As shown in Table \ref{tab:data:edit}, the final edit rates against MAPS are substantial, underscoring the necessity of this extensive human audit.

\section{Experiments}
With the benchmark tasks rigorously verified, we evaluated frontier LLMs on the resulting GAIA-v2-LILT dataset to measure the true impact of our review workflow.

\paragraph{Setup}
We evaluated three leading models (GPT-5.4, Gemini 3.1 Pro, Claude Opus 4.6) using the Open Deep Research agent \cite{smolagents}.
The agents were equipped with web search (Exa\footnote{\url{https://exa.ai}}), speech recognition, image captioning, and file read tools.
For each task, we limited the manager agent to 12 steps and the search subagent to 20 steps.

\begin{table*}[!ht]
  \centering
  \begin{tabular}{l c@{\;}c c@{\;}c c@{\;}c c@{\;}c c@{\;}c c}
    \toprule
    & \multicolumn{2}{c}{Arabic} & \multicolumn{2}{c}{German} & \multicolumn{2}{c}{Hindi} & \multicolumn{2}{c}{Korean} & \multicolumn{2}{c}{Portuguese} & \\
    \cmidrule(lr){2-3}\cmidrule(lr){4-5}\cmidrule(lr){6-7}\cmidrule(lr){8-9}\cmidrule(lr){10-11}
    Model & MT & Audit & MT & Audit & MT & Audit & MT & Audit & MT & Audit & English\\
    \midrule
    GPT-5.4 & 32.1 & 47.3 & 47.3 & 63.6 & 34.6 & 60.0 & 33.3 & 62.4 & 47.3 & 58.2 & 66.7 \\
    Gemini 3.1 Pro & 34.6 & 52.1 & 49.7 & 66.7 & 38.2 & 63.6 & 34.6 & 64.8 & 48.5 & 65.5 & 73.9 \\
    Claude Opus 4.6 & 32.1 & 49.1 & 49.7 & 66.7 & 29.7 & 62.4 & 33.3 & 58.8 & 49.1 & 63.0 & 79.4 \\
    \bottomrule
  \end{tabular}
  \caption{Agent performance before and after correction (pass@1 accuracy [\%]).}
  \label{tab:exp:perf}
\end{table*}

\paragraph{Impact of Audits}
Table \ref{tab:exp:perf} compares model performance on the raw MT drafts versus the human-corrected final data.
Before correction, multilingual scores lag substantially behind the English baselines.
However, after human audit, performance across all five target languages improves dramatically, yielding absolute gains ranging from +10.9\% to +32.7\%.
This recovery reduces the performance gap with English to a minimum of 3.1\%, though Arabic still shows a larger gap of up to 30.3\%.

These results demonstrate that MT may mask an agent's true capabilities significantly.
Our audits reveal the actual performance trends across languages, proving that carefully aligned data is essential for accurate multilingual benchmarking of agents.

\paragraph{Analysis of Edits}
To understand how the quality issues discussed in Section \ref{sec:multi} affect benchmark evaluation, we compare model outputs on the original and corrected versions of the same tasks.
For each task, we check whether linguists flagged a specific issue category during the audit.
We then measure whether correcting that issue caused the model's correctness to flip, changing the outcome from wrong to right or vice versa.

\begin{figure}[t]
  \centering
  \includegraphics[width=\columnwidth]{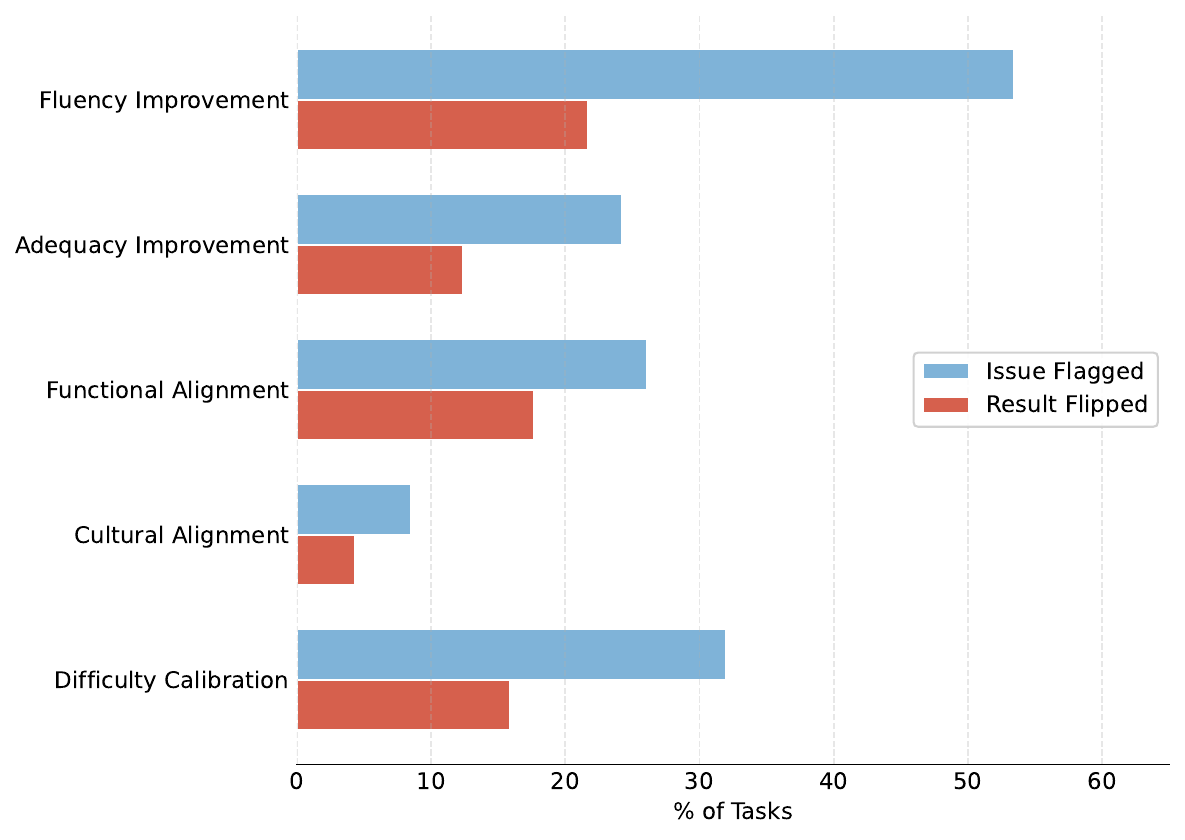}
  \caption{Flagging rates of issues during the audit and result flipping rates (Gemini 3.1 Pro) among the flagged tasks. Aggregated across all languages.}
  \label{fig:exp:flag}
\end{figure}

Figure \ref{fig:exp:flag} shows the analysis.
Since multiple issues can co-occur on the same task, flip rates are correlational rather than causal.
Nevertheless, functional alignment stands out with the highest flip rate (67.9\%), consistent with the expectation that a mistranslated gold answer penalizes the model regardless of actual performance.
Cultural alignment, adequacy, and difficulty calibration flip outcomes roughly half the time, indicating strong association with correctness judgments.
Notably, despite being the most frequently flagged, fluency issues have the lowest flip rate (40.6\%).

This suggests that stylistic perfection is weakly associated with evaluation outcome changes.
Consequently, resource-constrained localization efforts should prioritize functional and cultural fidelity to maximize the integrity of the evaluation.

\section{Conclusion}
Building on MAPS \cite{hofman2025maps}, we created the GAIA-v2-LILT dataset to directly address the task-logic failures caused by MT in agent benchmarks.
We developed a rigorous review workflow combining deterministic filters, single-axis LLM judges, and structured human audits to mitigate both LLM self-preference and human fluency bias.

By correcting functional and cultural misalignments while preserving task difficulty, models improved accuracy by up to 32.7\%.
This highlights the critical, yet often overlooked, impact of non-linguistic alignment errors when translating agent benchmarks.
We argue that future multilingual agent evaluations must adopt similar workflows to ensure accurate measurement.

\section*{Limitations}
Our review process omits baseline agent testing which can detect hidden technical flaws, e.g., search APIs failing to retrieve required information (due to poor retrieval or geo-blocking) or models bypassing intended tool usage by relying on memorized training data.
Since configuring execution environments is impractical for non-technical annotators, we leave the integration of automated agent testing into the review loop via annotator-friendly interfaces to future work.

Also, file attachments (e.g., images, audio, and PDFs) were not localized, as building a consistent editing pipeline for diverse asset types is highly complex.
While keeping these files in English does not break the core task logic, it prevents perfect localization.
The development of streamlined multimodal localization workflows is left to future work.

\section*{Acknowledgments}
We sincerely thank the human annotators for reviewing the GAIA-v2-LILT dataset, and the authors of MAPS \cite{hofman2025maps} for coordinating the data release.

\bibliography{custom}

@article{mialon2023augmented,
title={Augmented Language Models: a Survey},
author={Gr{\'e}goire Mialon and Roberto Dessi and Maria Lomeli and Christoforos Nalmpantis and Ramakanth Pasunuru and Roberta Raileanu and Baptiste Roziere and Timo Schick and Jane Dwivedi-Yu and Asli Celikyilmaz and Edouard Grave and Yann LeCun and Thomas Scialom},
journal={Transactions on Machine Learning Research},
issn={2835-8856},
year={2023},
url={https://openreview.net/forum?id=jh7wH2AzKK},
}

@article{wang2024autonomousagentssurvey,
  title   = {A Survey on Large Language Model Based Autonomous Agents},
  author  = {Wang, Lei and Ma, Chen and Feng, Xueyang and Zhang, Zeyu and Yang, Hao and Zhang, Jingsen and Chen, Zhiyuan and Tang, Jiakai and Chen, Xu and Lin, Yankai and Zhao, Wayne Xin and Wei, Zhewei and Wen, Jirong},
  journal = {Frontiers of Computer Science},
  volume  = {18},
  pages   = {186345},
  year    = {2024},
  doi     = {10.1007/s11704-024-40231-1},
  url     = {https://link.springer.com/article/10.1007/s11704-024-40231-1}
}

@article{xi2025risepotential,
  title   = {The Rise and Potential of Large Language Model Based Agents: A Survey},
  author  = {Xi, Zhiheng and Chen, Wenxiang and Guo, Xin and He, Wei and Ding, Yiwen and Hong, Boyang and Zhang, Ming and Wang, Junzhe and Jin, Senjie and Zhou, Enyu and Zheng, Rui and Fan, Xiaoran and Wang, Xiao and Xiong, Limao and Zhou, Yuhao and Wang, Weiran and Jiang, Changhao and Zou, Yicheng and Liu, Xiangyang and Yin, Zhangyue and Dou, Shihan and Weng, Rongxiang and Qin, Wenjuan and Zheng, Yongyan and Qiu, Xipeng and Huang, Xuanjing and Zhang, Qi and Gui, Tao},
  journal = {Science China Information Sciences},
  volume  = {68},
  number  = {2},
  pages   = {121101},
  year    = {2025},
  doi     = {10.1007/s11432-024-4222-0},
  url     = {https://link.springer.com/article/10.1007/s11432-024-4222-0}
}

@inproceedings{yao2022webshop,
  title     = {WebShop: Towards Scalable Real-World Web Interaction with Grounded Language Agents},
  author    = {Yao, Shunyu and Chen, Howard and Yang, John and Narasimhan, Karthik},
  booktitle = {Advances in Neural Information Processing Systems},
  volume    = {35},
  pages     = {20744--20757},
  publisher = {Curran Associates, Inc.},
  year      = {2022},
  note      = {NeurIPS 2022 Main Conference Track},
  url       = {https://proceedings.neurips.cc/paper_files/paper/2022/hash/82ad13ec01f9fe44c01cb91814fd7b8c-Abstract-Conference.html}
}

@inproceedings{deng2023mind2web,
  title     = {Mind2Web: Towards a Generalist Agent for the Web},
  author    = {Deng, Xiang and Gu, Yu and Zheng, Boyuan and Chen, Shijie and Stevens, Samuel and Wang, Boshi and Sun, Huan and Su, Yu},
  booktitle = {Advances in Neural Information Processing Systems},
  volume    = {36},
  pages     = {28091--28114},
  publisher = {Curran Associates, Inc.},
  year      = {2023},
  note      = {NeurIPS 2023 Datasets and Benchmarks Track},
  url       = {https://papers.nips.cc/paper_files/paper/2023/hash/5950bf290a1570ea401bf98882128160-Abstract-Datasets_and_Benchmarks.html}
}

@inproceedings{liu2024agentbench,
  title     = {AgentBench: Evaluating {LLM}s as Agents},
  author    = {Liu, Xiao and Yu, Hao and Zhang, Hanchen and Xu, Yifan and Lei, Xuanyu and Lai, Hanyu and Gu, Yu and Ding, Hangliang and Men, Kaiwen and Yang, Kejuan and Zhang, Shudan and Deng, Xiang and Zeng, Aohan and Du, Zhengxiao and Zhang, Chenhui and Shen, Sheng and Zhang, Tianjun and Su, Yu and Sun, Huan and Huang, Minlie and Dong, Yuxiao and Tang, Jie},
  booktitle = {The Twelfth International Conference on Learning Representations},
  year      = {2024},
  note      = {Poster},
  url       = {https://openreview.net/forum?id=zAdUB0aCTQ}
}

@inproceedings{zhou2024webarena,
  title     = {WebArena: A Realistic Web Environment for Building Autonomous Agents},
  author    = {Zhou, Shuyan and Xu, Frank F. and Zhu, Hao and Zhou, Xuhui and Lo, Robert and Sridhar, Abishek and Cheng, Xianyi and Ou, Tianyue and Bisk, Yonatan and Fried, Daniel and Alon, Uri and Neubig, Graham},
  booktitle = {The Twelfth International Conference on Learning Representations},
  year      = {2024},
  note      = {Poster},
  url       = {https://openreview.net/forum?id=oKn9c6ytLx}
}

@inproceedings{jimenez2024swebench,
  title     = {{SWE}-bench: Can Language Models Resolve Real-World {GitHub} Issues?},
  author    = {Jimenez, Carlos E. and Yang, John and Wettig, Alexander and Yao, Shunyu and Pei, Kexin and Press, Ofir and Narasimhan, Karthik R.},
  booktitle = {The Twelfth International Conference on Learning Representations},
  year      = {2024},
  note      = {Oral},
  url       = {https://openreview.net/forum?id=VTF8yNQM66}
}

@article{mialon2023gaia,
  title   = {GAIA: a benchmark for General AI Assistants},
  author  = {Mialon, Gr{\'e}goire and Fourrier, Cl{\'e}mentine and Swift, Craig and Wolf, Thomas and LeCun, Yann and Scialom, Thomas},
  journal = {arXiv preprint arXiv:2311.12983},
  year    = {2023},
  doi     = {10.48550/arXiv.2311.12983},
  url     = {https://arxiv.org/abs/2311.12983}
}

@article{barres2025tau2bench,
  title   = {{$\tau^2$-Bench: Evaluating Conversational Agents in a Dual-Control Environment}},
  author  = {Barres, Victor and Dong, Honghua and Ray, Soham and Si, Xujie and Narasimhan, Karthik},
  journal = {arXiv preprint arXiv:2506.07982},
  year    = {2025},
  doi     = {10.48550/arXiv.2506.07982},
  url     = {https://arxiv.org/abs/2506.07982}
}

@article{wei2025browsecomp,
  title   = {BrowseComp: A Simple Yet Challenging Benchmark for Browsing Agents},
  author  = {Wei, Jason and Sun, Zhiqing and Papay, Spencer and McKinney, Scott and Han, Jeffrey and Fulford, Isa and Chung, Hyung Won and Passos, Alex Tachard and Fedus, William and Glaese, Amelia},
  journal = {arXiv preprint arXiv:2504.12516},
  year    = {2025},
  doi     = {10.48550/arXiv.2504.12516},
  url     = {https://arxiv.org/abs/2504.12516}
}

@inproceedings{xie2024osworld,
  title     = {OSWorld: Benchmarking Multimodal Agents for Open-Ended Tasks in Real Computer Environments},
  author    = {Xie, Tianbao and Zhang, Danyang and Chen, Jixuan and Li, Xiaochuan and Zhao, Siheng and Cao, Ruisheng and Hua, Toh Jing and Cheng, Zhoujun and Shin, Dongchan and Lei, Fangyu and Liu, Yitao and Xu, Yiheng and Zhou, Shuyan and Savarese, Silvio and Xiong, Caiming and Zhong, Victor and Yu, Tao},
  booktitle = {Advances in Neural Information Processing Systems},
  volume    = {37},
  pages     = {52040--52094},
  publisher = {Curran Associates, Inc.},
  year      = {2024},
  note      = {NeurIPS 2024 Datasets and Benchmarks Track},
  url       = {https://proceedings.neurips.cc/paper_files/paper/2024/hash/5d413e48f84dc61244b6be550f1cd8f5-Abstract-Datasets_and_Benchmarks_Track.html}
}

@article{patwardhan2025gdpval,
  title   = {GDPval: Evaluating AI Model Performance on Real-World Economically Valuable Tasks},
  author  = {Patwardhan, Tejal and Dias, Rachel and Proehl, Elizabeth and Kim, Grace and Wang, Michele and Watkins, Olivia and Posada Fishman, Sim{\'o}n and Aljubeh, Marwan and Thacker, Phoebe and Fauconnet, Laurance and Kim, Natalie S. and Chao, Patrick and Miserendino, Samuel and Chabot, Gildas and Li, David and Sharman, Michael and Barr, Alexandra and Glaese, Amelia and Tworek, Jerry},
  journal = {arXiv preprint arXiv:2510.04374},
  year    = {2025},
  doi     = {10.48550/arXiv.2510.04374},
  url     = {https://arxiv.org/abs/2510.04374}
}

@inproceedings{ahuja2023mega,
  title     = {{MEGA}: Multilingual Evaluation of Generative {AI}},
  author    = {Ahuja, Kabir and Diddee, Harshita and Hada, Rishav and Ochieng, Millicent and Ramesh, Krithika and Jain, Prachi and Nambi, Akshay and Ganu, Tanuja and Segal, Sameer and Ahmed, Mohamed and Bali, Kalika and Sitaram, Sunayana},
  booktitle = {Proceedings of the 2023 Conference on Empirical Methods in Natural Language Processing},
  address   = {Singapore},
  publisher = {Association for Computational Linguistics},
  pages     = {4232--4267},
  year      = {2023},
  doi       = {10.18653/v1/2023.emnlp-main.258},
  url       = {https://aclanthology.org/2023.emnlp-main.258/}
}

@inproceedings{zhang2023donttrust,
  title     = {Don{'}t Trust {ChatGPT} when your Question is not in English: A Study of Multilingual Abilities and Types of {LLM}s},
  author    = {Zhang, Xiang and Li, Senyu and Hauer, Bradley and Shi, Ning and Kondrak, Grzegorz},
  booktitle = {Proceedings of the 2023 Conference on Empirical Methods in Natural Language Processing},
  address   = {Singapore},
  publisher = {Association for Computational Linguistics},
  pages     = {7915--7927},
  year      = {2023},
  doi       = {10.18653/v1/2023.emnlp-main.491},
  url       = {https://aclanthology.org/2023.emnlp-main.491/}
}

@inproceedings{nguyen2024culturax,
  title     = {{CulturaX}: A Cleaned, Enormous, and Multilingual Dataset for Large Language Models in 167 Languages},
  author    = {Nguyen, Thuat and Nguyen, Chien Van and Lai, Viet Dac and Man, Hieu and Ngo, Nghia Trung and Dernoncourt, Franck and Rossi, Ryan A. and Nguyen, Thien Huu},
  booktitle = {Proceedings of the 2024 Joint International Conference on Computational Linguistics, Language Resources and Evaluation ({LREC}-{COLING} 2024)},
  address   = {Torino, Italia},
  publisher = {ELRA and ICCL},
  pages     = {4226--4237},
  year      = {2024},
  url       = {https://aclanthology.org/2024.lrec-main.377/}
}

@article{hofman2025maps,
  title   = {MAPS: A Multilingual Benchmark for Agent Performance and Security},
  author  = {Hofman, Omer and Brokman, Jonathan and Rachmil, Oren and Bose, Shamik and Pahuja, Vikas and Shimizu, Toshiya and Starostina, Trisha and Marchisio, Kelly and Goldfarb-Tarrant, Seraphina and Vainshtein, Roman},
  journal = {arXiv preprint arXiv:2505.15935},
  year    = {2025},
  doi     = {10.48550/arXiv.2505.15935},
  url     = {https://arxiv.org/abs/2505.15935},
  note    = {Accepted to Findings of EACL 2026}
}

@inproceedings{wang2025xwebagentbench,
  title     = {X-WebAgentBench: Evaluating Multilingual LLM Agents on Multimodal Interactive Web Navigation Tasks},
  author    = {Wang, Peng and Tao, Chunping and Niu, Lu and Jiang, Hao and Qiao, Shuai and Xiang, Ji and Han, Jiawei and Xu, Haoran and Xu, Chao and Ge, Tairan and Tu, Zhaopeng and Zhao, Wayne Xin and Wen, Ji-Rong},
  booktitle = {Findings of the Association for Computational Linguistics: ACL 2025},
  address   = {Vienna, Austria},
  publisher = {Association for Computational Linguistics},
  pages     = {19097--19122},
  year      = {2025},
  doi       = {10.18653/v1/2025.findings-acl.988},
  url       = {https://aclanthology.org/2025.findings-acl.988/}
}

@article{issaka2026translationproxy,
  title   = {Translation as a Scalable Proxy for Multilingual Evaluation},
  author  = {Issaka, Sheriff and Rosas Gonzalez, Erick and Liu, Lieqi and Agyei, Evans Kofi and Bandarkar, Lucas and Peng, Nanyun and Adelani, David Ifeoluwa and Guzm{\'a}n, Francisco and Gabriel, Saadia},
  journal = {arXiv preprint arXiv:2601.11778},
  year    = {2026},
  doi     = {10.48550/arXiv.2601.11778},
  url     = {https://arxiv.org/abs/2601.11778}
}

@article{yang2025macosworld,
  title   = {{macOSWorld}: A Multilingual Interactive Benchmark for {GUI} Agents},
  author  = {Yang, Pei and Ci, Hai and Shou, Mike Zheng},
  journal = {arXiv preprint arXiv:2506.04135},
  year    = {2025},
  doi     = {10.48550/arXiv.2506.04135},
  url     = {https://arxiv.org/abs/2506.04135}
}

@inproceedings{kulkarni2025massiveagents,
  title     = {{MASSIVE}-Agents: A Benchmark for Multilingual Function Calling in 52 Languages},
  author    = {Kulkarni, Vaibhav and Mhetre, Tushar and Singh, Amit and Saha, Ripan and Chatterjee, Saurav and Raj, Sreevijay and Swamy, Sandesh and Dave, Shachi and Srinivasan, Raghavan and Das, Diptanu and Govindaraju, Venu},
  booktitle = {Proceedings of the 2025 Conference on Empirical Methods in Natural Language Processing},
  address   = {Suzhou, China},
  publisher = {Association for Computational Linguistics},
  pages     = {9309--9342},
  year      = {2025},
  doi       = {10.18653/v1/2025.emnlp-main.513},
  url       = {https://aclanthology.org/2025.emnlp-main.513/}
}

@inproceedings{raihan2025mhumaneval,
  title={mHumanEval-a multilingual benchmark to evaluate large language models for code generation},
  author={Raihan, Md Nishat and Anastasopoulos, Antonios and Zampieri, Marcos},
  booktitle={Proceedings of the 2025 Conference of the Nations of the Americas Chapter of the Association for Computational Linguistics: Human Language Technologies (Volume 1: Long Papers)},
  pages={11432--11461},
  year={2025}
}

@article{luo2026lost,
  title={Lost in Execution: On the Multilingual Robustness of Tool Calling in Large Language Models},
  author={Luo, Zheng and Kutralingam, T Pranav and Okoani, Ogochukwu N and Xu, Wanpeng and Wei, Hua and Hu, Xiyang},
  journal={arXiv preprint arXiv:2601.05366},
  year={2026}
}

@article{koponen2016machine,
  title={Is machine translation post-editing worth the effort? A survey of research into post-editing and effort},
  author={Koponen, Maarit},
  journal={The Journal of Specialised Translation},
  number={25},
  pages={131--148},
  year={2016}
}

@incollection{vieira2019post,
  title={Post-editing of machine translation},
  author={Vieira, Lucas Nunes},
  booktitle={The Routledge handbook of translation and technology},
  pages={319--336},
  year={2019},
  publisher={Routledge}
}

@book{balling2014post,
  title={Post-editing of machine translation: Processes and applications},
  author={Balling, Laura Winther and Carl, Michael and O'Brian, Sharon},
  year={2014},
  publisher={Cambridge Scholars Publishing}
}

@inproceedings{lommel2013multidimensional,
  title={Multidimensional quality metrics: a flexible system for assessing translation quality},
  author={Lommel, Arle Richard and Burchardt, Aljoscha and Uszkoreit, Hans},
  booktitle={Proceedings of Translating and the Computer 35},
  year={2013}
}

@article{castilho2025survey,
  title={A survey of context in neural machine translation and its evaluation},
  author={Castilho, Sheila and Knowles, Rebecca},
  journal={Natural Language Processing},
  volume={31},
  number={4},
  pages={986--1016},
  year={2025},
  publisher={Cambridge University Press}
}

@article{agrawal2024assessing,
  title={Assessing the Role of Context in Chat Translation Evaluation: Is Context Helpful and Under What Conditions?},
  author={Agrawal, Sweta and Farajian, Amin and Fernandes, Patrick and Rei, Ricardo and Martins, Andr{\'e} FT},
  journal={Transactions of the Association for Computational Linguistics},
  volume={12},
  pages={1250--1267},
  year={2024},
  publisher={MIT Press}
}

@inproceedings{koneru2024contextual,
  title={Contextual refinement of translations: Large language models for sentence and document-level post-editing},
  author={Koneru, Sai and Exel, Miriam and Huck, Matthias and Niehues, Jan},
  booktitle={Proceedings of the 2024 Conference of the North American Chapter of the Association for Computational Linguistics: Human Language Technologies (Volume 1: Long Papers)},
  pages={2711--2725},
  year={2024}
}

@article{almeida2025ticket,
  title={Ticket-Bench: A Kickoff for Multilingual and Regionalized Agent Evaluation},
  author={Almeida, Thales Sales and Santos, Jo{\~a}o Guilherme Alves and Laitz, Thiago and Bon{\'a}s, Giovana Kerche},
  journal={arXiv preprint arXiv:2509.14477},
  year={2025}
}

@article{kautsar2025seadialogues,
  title={SEADialogues: A Multilingual Culturally Grounded Multi-turn Dialogue Dataset on Southeast Asian Languages},
  author={Kautsar, Muhammad Dehan Al and Candra, Aswin and Hakim, Muhammad Alif Al and Kahfi, Maxalmina Satria and Koto, Fajri and Aji, Alham Fikri and Limkonchotiwat, Peerat and Chuangsuwanich, Ekapol and Winata, Genta Indra},
  journal={arXiv preprint arXiv:2508.07069},
  year={2025}
}

@article{zhou2025browsecomp,
  title={Browsecomp-zh: Benchmarking web browsing ability of large language models in chinese},
  author={Zhou, Peilin and Leon, Bruce and Ying, Xiang and Zhang, Can and Shao, Yifan and Ye, Qichen and Chong, Dading and Jin, Zhiling and Xie, Chenxuan and Cao, Meng and others},
  journal={arXiv preprint arXiv:2504.19314},
  year={2025}
}

@inproceedings{koppel2011translationese,
  title={Translationese and its dialects},
  author={Koppel, Moshe and Ordan, Noam},
  booktitle={Proceedings of the 49th annual meeting of the association for computational linguistics: Human language technologies},
  pages={1318--1326},
  year={2011}
}

@inproceedings{li2025lost,
  title={Lost in literalism: How supervised training shapes translationese in LLMs},
  author={Li, Yafu and Zhang, Ronghao and Wang, Zhilin and Zhang, Huajian and Cui, Leyang and Yin, Yongjing and Xiao, Tong and Zhang, Yue},
  booktitle={Proceedings of the 63rd Annual Meeting of the Association for Computational Linguistics (Volume 1: Long Papers)},
  pages={12875--12894},
  year={2025}
}

@article{huang2025survey,
  title={A survey on hallucination in large language models: Principles, taxonomy, challenges, and open questions},
  author={Huang, Lei and Yu, Weijiang and Ma, Weitao and Zhong, Weihong and Feng, Zhangyin and Wang, Haotian and Chen, Qianglong and Peng, Weihua and Feng, Xiaocheng and Qin, Bing and others},
  journal={ACM Transactions on Information Systems},
  volume={43},
  number={2},
  pages={1--55},
  year={2025},
  publisher={ACM New York, NY}
}

@article{zheng2023judging,
  title={Judging llm-as-a-judge with mt-bench and chatbot arena},
  author={Zheng, Lianmin and Chiang, Wei-Lin and Sheng, Ying and Zhuang, Siyuan and Wu, Zhanghao and Zhuang, Yonghao and Lin, Zi and Li, Zhuohan and Li, Dacheng and Xing, Eric and others},
  journal={Advances in neural information processing systems},
  volume={36},
  pages={46595--46623},
  year={2023}
}

@inproceedings{gudibande2024false,
  title={The false promise of imitating proprietary language models},
  author={Gudibande, Arnav and Wallace, Eric and Snell, Charlie Victor and Geng, Xinyang and Liu, Hao and Abbeel, Pieter and Levine, Sergey and Song, Dawn},
  booktitle={The Twelfth International Conference on Learning Representations},
  year={2024}
}

@inproceedings{li2025language,
  title={Language ranker: A metric for quantifying llm performance across high and low-resource languages},
  author={Li, Zihao and Shi, Yucheng and Liu, Zirui and Yang, Fan and Payani, Ali and Liu, Ninghao and Du, Mengnan},
  booktitle={Proceedings of the AAAI Conference on Artificial Intelligence},
  volume={39},
  number={27},
  pages={28186--28194},
  year={2025}
}

@inproceedings{wang2025lost,
  title={Lost in multilinguality: Dissecting cross-lingual factual inconsistency in transformer language models},
  author={Wang, Mingyang and Adel, Heike and Lange, Lukas and Liu, Yihong and Nie, Ercong and Str{\"o}tgen, Jannik and Sch{\"u}tze, Hinrich},
  booktitle={Proceedings of the 63rd Annual Meeting of the Association for Computational Linguistics (Volume 1: Long Papers)},
  pages={5075--5094},
  year={2025}
}

@inproceedings{saha2024branch,
  title={Branch-solve-merge improves large language model evaluation and generation},
  author={Saha, Swarnadeep and Levy, Omer and Celikyilmaz, Asli and Bansal, Mohit and Weston, Jason and Li, Xian},
  booktitle={Proceedings of the 2024 Conference of the North American Chapter of the Association for Computational Linguistics: Human Language Technologies (Volume 1: Long Papers)},
  pages={8352--8370},
  year={2024}
}

@inproceedings{feng2025m,
  title={M-MAD: Multidimensional multi-agent debate for advanced machine translation evaluation},
  author={Feng, Zhaopeng and Su, Jiayuan and Zheng, Jiamei and Ren, Jiahan and Zhang, Yan and Wu, Jian and Wang, Hongwei and Liu, Zuozhu},
  booktitle={Proceedings of the 63rd Annual Meeting of the Association for Computational Linguistics (Volume 1: Long Papers)},
  pages={7084--7107},
  year={2025}
}

@inproceedings{lee2025checkeval,
  title={Checkeval: A reliable llm-as-a-judge framework for evaluating text generation using checklists},
  author={Lee, Yukyung and Kim, Joonghoon and Kim, Jaehee and Cho, Hyowon and Kang, Jaewook and Kang, Pilsung and Kim, Najoung},
  booktitle={Proceedings of the 2025 Conference on Empirical Methods in Natural Language Processing},
  pages={15782--15809},
  year={2025}
}

@Misc{smolagents,
  title =        {`smolagents`: a smol library to build great agentic systems},
  author =       {Aymeric Roucher and Albert Villanova del Moral and Thomas Wolf and Leandro von Werra and Erik Kaunismäki},
  howpublished = {\url{https://github.com/huggingface/smolagents}},
  year =         {2025}
}

\end{document}